\documentclass[11pt,a4paper]{article}
\usepackage[hyperref]{emnlp2019}
\usepackage{times}
\usepackage{latexsym}
\usepackage[caption = false]{subfig}
\usepackage[final]{graphicx}
\usepackage{amsmath,amsthm,amssymb}
\usepackage{multirow}
\usepackage{makecell}

\usepackage{url}

\aclfinalcopy 

\usepackage{booktabs}

\DeclareMathOperator*{\embed}{embed}
\DeclareMathOperator*{\lstm}{LSTM}
\DeclareMathOperator*{\softmax}{softmax}
\DeclareMathOperator*{\mlp}{MLP}

\usepackage{cleveref}
\usepackage{float}

\title{Comparison of Interactive Knowledge Base Spelling Correction Models for Low-Resource Languages}

\author{Yiyuan Li$^1$, Antonios Anastasopoulos$^2$, Alan W Black$^2$\\
  $^1$Department of Electrical and Computer Engineering \\
  $^2$Language Technology Institute\\
  Carnegie Mellon University \\
  {\tt yiyuanli@andrew.cmu.edu aanastas@cs.cmu.edu awb@cs.cmu.edu}}

\date{}

\begin{document}
\maketitle

\begin{abstract}
Spelling normalization for low resource languages is a challenging task because the patterns are hard to predict and large corpora are usually required to collect enough examples. This work shows a comparison of a neural model and character language models with varying amounts on target language data.  Our usage scenario is interactive correction with nearly zero amounts of training examples, improving models as more data is collected, for example within a chat app.  Such models are designed to be incrementally improved as feedback is given from users. In this work, we design a knowledge-base and prediction model embedded system for spelling correction in low-resource languages. Experimental results on multiple languages show that the model could become effective with a small amount of data. We perform experiments on both natural and synthetic data, as well as on data from two endangered languages (Ainu and Griko).
Last, we built a prototype system that was used for a small case study on Hinglish, which further demonstrated the suitability of our approach in real world scenarios.
\end{abstract}

\section{Introduction}
Online writing in high resource languages is now normally accompanied by spelling correction software but such a relatively simple technology is rarely available for very low resource languages.  Although it is well defined how to build such spelling correction models, it typically requires a large amount of data in the target language to build such models.

Low resource languages often are languages of low literacy (even if the speakers are literate in some other languages).  Low resource languages often also do not have the advance of years of standardization in spelling, and a long-term education system to spread this to speakers.  Encouraging literacy in low resource languages is typically good for language preservation.  Encouraging spelling standardization is also good for readability of the language for users, and also for further language technologies such as translation, grammar correction, etc.  Thus we wish to offer models that can follow a stream of written text in a language and make useful suggestions for standardization of the written forms with little or even no pre-training. 

\begin{figure}[t]
  \centering
  \includegraphics[width=0.9\columnwidth]{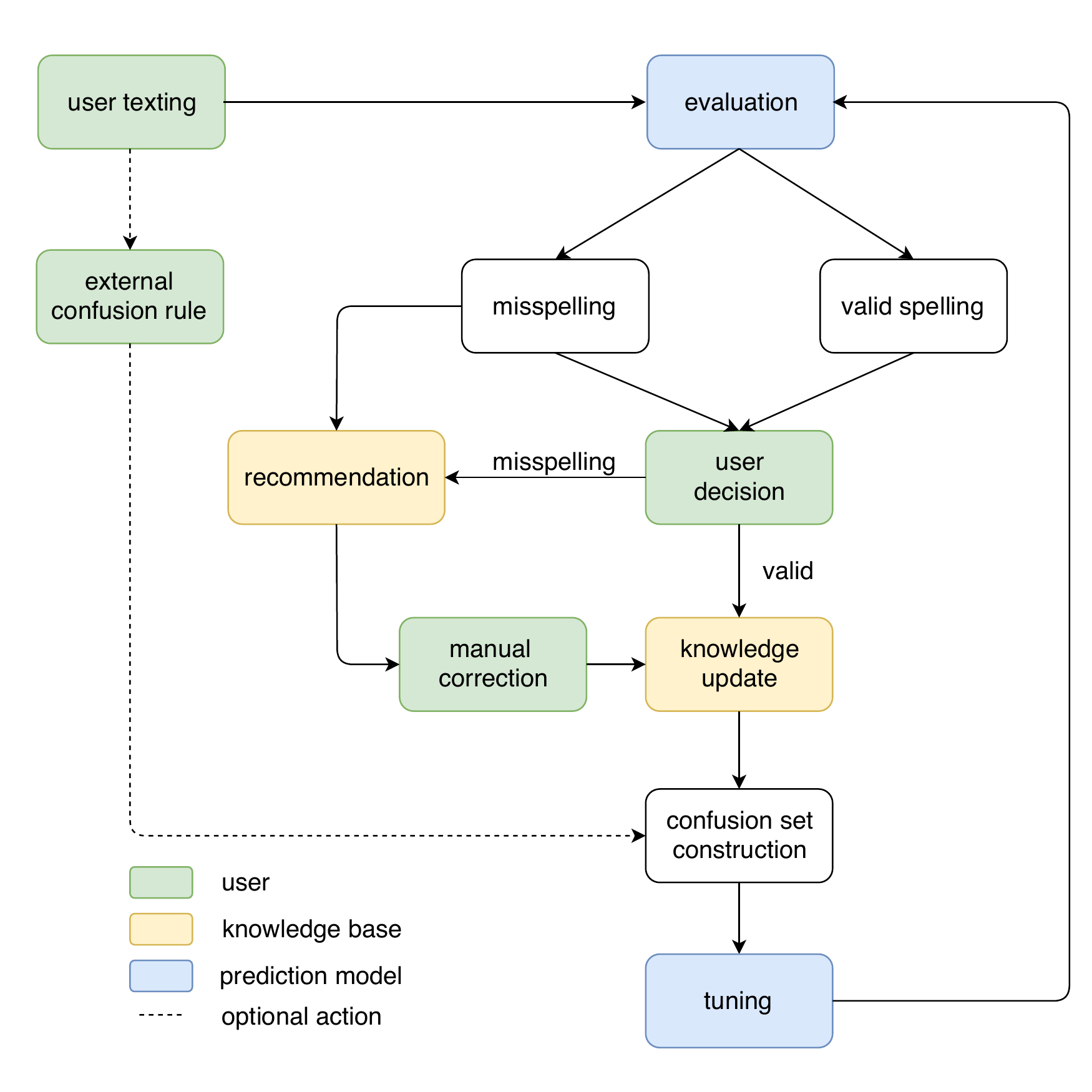} 
  \caption{Overview of our system design. The user, knowledge base and prediction model interact with each other, represented by green, yellow and blue blocks respectively.}
  \label{fig: structure} 
\end{figure}

Spelling normalization is a task to identify misspelled words from text sequences. For example, English speakers may misspell \textit{acquit} to \textit{aquit} and \textit{address} to \textit{adress}. Misspelling could be divided into real word error and non-word error~\cite{Chen2007ImprovingQS}. The former means incorrect usage of valid words (e.g. \textit{adapt} and \textit{adopt}) or variants of same word under different social settings (e.g. \textit{center} and \textit{centre}, which are common spellings in American and British English respectively), the latter leads to non-existent words which cause misunderstanding. It is hard to identify global misspelling patterns since it relates to various cultural factors and input environments like keyboard design. Golden standards like dictionaries are not flexible to grow with newly-generated and unpredictable forms of the vocabulary. Additionally, a dictionary could only prove correctness of words but not vice versa. Thus, it requires common sense and morphological knowledge to successfully identify misspellings. The task for machine 
learning is to have a sufficiently general model that can recognize if a novel token is misspelled or is simply a new word. Also, for many low-resource languages, neither a well-formed dictionary nor spelling checkers built on enough misspelling samples is accessible.  Therefore, it is important accurately make decisions about the misspellings and adopt new misspellings from a limited corpus.

Since misspellings only rarely occur, it leads to an extreme imbalance of labels in the existing corpus when formulated into a classification problem. Moreover, it is hard to cover morphological features in the misspellings with limited training data for some morphological-rich languages like Finnish. Existing methods to leverage unbalanced classes like cost-sensitive learning ~\cite{Kukar1998CostSensitiveLW} and sampling may not be appropriate. The former relies on fixed cost matrix and is negative to scalability, and the latter reduces the amount of data in this resource-sensitive scenario. Therefore, we attempt to identify spelling patterns with only small set of almost-correct tokens. We employ a random-pattern mixed strategy to augment the training data. 

In this paper, we try to capture the spelling patterns through an interactive method with a small set of seed words ($\sim$1000). We propose a knowledge-base and classifier combined model to identify misspellings. A data augmentation strategy is employed to construct a confusion set of synthetic misspellings.  The knowledge base contains the list of known correct words, either from pre-existing data, or updated as the user interactively accepts identified misspelled words as actually correct.  

The main contribution of this work is a method in mitigating rareness of misspellings and the system we develop that actively boost performance from the interactions with human users. We conduct experiments under three settings. First, we focus on real spelling mistakes based on available English and Russian corpora. Second, we experiment with plausible synthetic spelling mistakes in English, Spanish, Turkish, and Finish. 
Lastly, we conduct a small-scale real-world case study in Hinglish (romanized Hindi+English), where we find that our system to be effective even with only a few hundred words. 

\section{Related Work}

Spelling normalization is sub-task of spelling correction~\cite{kukich1992techniques}, which being an important task for text processing and normalization~\cite{Gong2019ContextSensitiveMS}, and could be considered as a type of grammar correction~\cite{ng-etal-2014-conll}. Previous spelling correction schemes include edit distance~\cite{Ristad1996LearningSE}, noisy channel model~\cite{nagata-1996-context}, and neural models~\cite{li2018spelling}. ~\citet{Sakaguchi2016RobsutWR} proposed a robust neural model for spelling correction, but needs a much larger corpus while could only deal with spelling errors of specific categories.~\citet{Pruthi2019CombatingAM} improved this model to be resilient to adversarial attacks by conducting more expensive training. However, their schemes are not close to the way lexicon varies in different culture environments.

Spelling correction is also widely applied in web query search, completion and suggestion~\cite{Chen2007ImprovingQS}, where interaction with users is embedded in the model such as noisy channels~\cite{Duan2011OnlineSC} and graphical models~\cite{li2017exploring}. However, click log or other external resource is required to assist the correction, which is not feasible in the low-resource setting. 

Note although we are envisaging an interactive learning system where user feedback is used to regularly update the models, this is not the same as active learning~\cite{Cohn1994}, as we have now pre-existing set of candidates to choose from to offer to the user for labeling.  

Our work differs from active learning in the following perspectives. First, unlabelled data is also expensive to access for low-resource languages, and we do not have a fixed pool of unlabelled training data or access of the full corpora to pick new unlabelled samples since we actively receive information from user. Rather, we ask the user to provide suggestion on the retrieved candidates decided by the model. In addition, our decision is made by asking human users for manual correction rather than the machine learning classifiers.

\section{System}

Importantly we care about the end user, so in addition to testing different models' performance of predicting correct/misspelling, we also care about how such models are involved in an end-user system.  Thus we built a simple instance of a typical system.  

An overview of the system is shown in Figure~\ref{fig: structure}, where we represent the three main sections of our system (user, knowledge base and prediction model) and how they interact with each other. The prediction model makes a decision on each word received from the user. Then the decision is delivered back for human re-evaluation. If the user accepts a misspelling decision, our system retrieves a candidates list of potential correct words from the knowledge base, from which the user can choose a correction or manually provide the correct form as an alternative. The knowledge base is consequently updated, and the new labelled pair is used to update prediction model. Details of the system is discussed as follows: Section~\ref{sec: UI} provides an introduction to the user interface and how it receives incremental information from the user; section~\ref{sec: knowledge base} demonstrates knowledge base which preserves information of existing correct spellings, and the prediction model employed is covered in section~\ref{sec: prediction model}.

\begin{figure}[t]
  \centering
  \includegraphics[width=1.1\columnwidth]{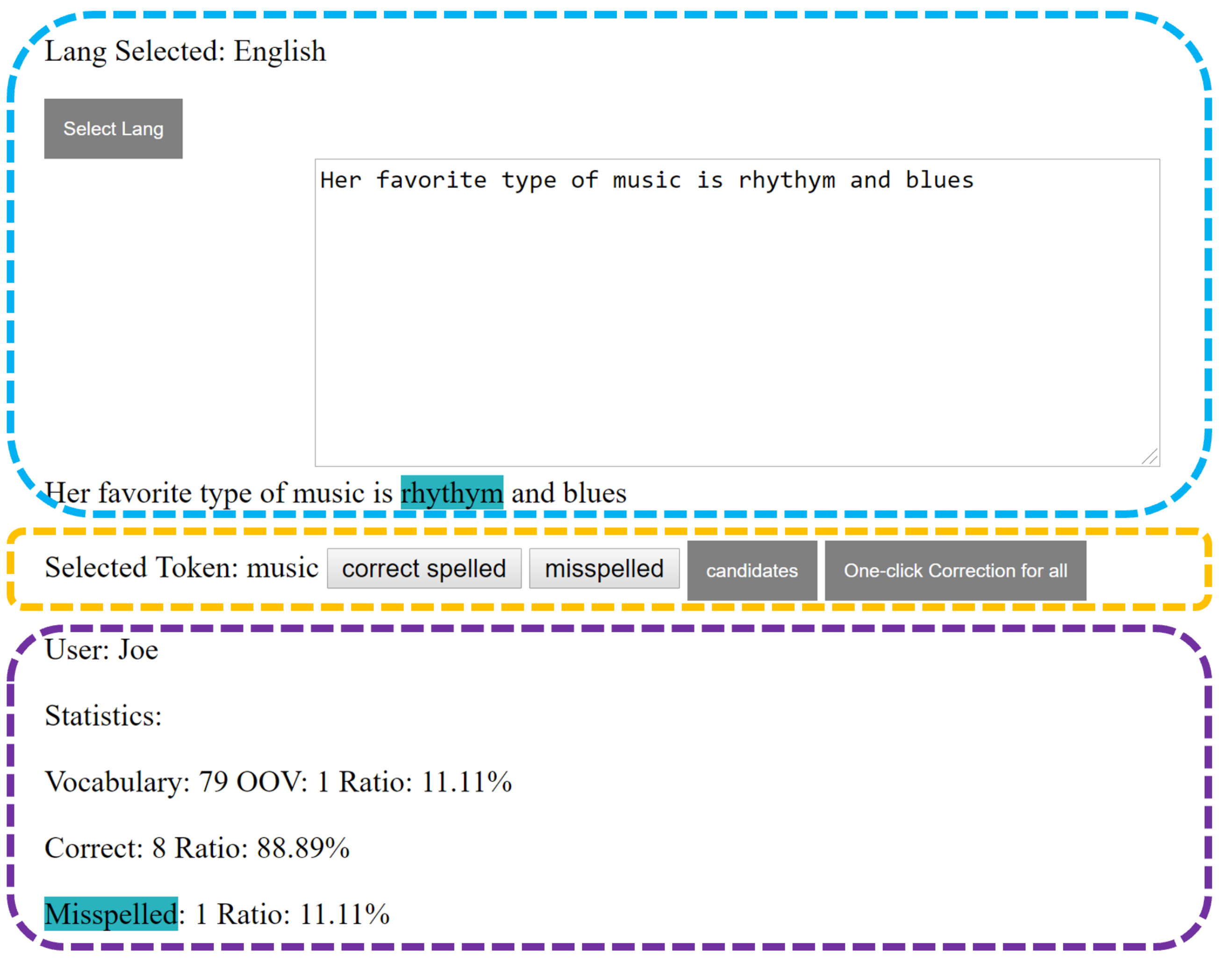} 
  \caption{Our user interface prototype. Users select language and can type in the blue area; they provide labels/corrections in the orange area, while summary statistics are displayed in the purple area.}
  \label{fig: UI} 
\end{figure}

\subsection{User Interface}
\label{sec: UI}
We developed an user interface to provide both a usable spelling correction systems with new languages, shown in Figure~\ref{fig: UI}. The user first needs to choose the language before typing, or name a new language. While typing in the text box, the completed text is simultaneously analyzed and dynamically displayed in the area below. Behind the scene, each token is delivered to the back-end prediction model and its prediction (correct/misspelled) is shown by the background color of the corresponding token block. The user can manually correct the decision of the model by selecting the token block and either make a correct spelling or misspelling decision. The former action updates the token to the knowledge base directly, while the latter action can also trigger a knowledge update by selecting one of the recommended spellings from the candidate list, or by having the user provide the correct spelling manually.

The prediction model collects the extra word with spelling label through both approaches and tunes the prediction model with the incremental information received at this round. The interface also supports text file uploading for direct corpus upload and customize-able training with labelled lexicon.

\subsection{Knowledge Base}
\label{sec: knowledge base}
The model employs a knowledge base to map all words it receives to their correct forms. While the prediction model is used to label unseen words, the user is involved to provide advice on whether that decision is correct or not. The knowledge base is updated in the following scheme: once the user types in a new word, the decision is made by the offline model and the knowledge base simultaneously. The model prediction will only be effective if the word is out-of-vocabulary, otherwise the system directly returns the result from the knowledge base whether this is a misspelling or not. 

In the case of a predicted misspelling, the system provides suggestions based on the current entries of the knowledge base. We measure the similarity of the misspelled word to the words in the knowledge base using Levenshtein edit distance~\cite{levenshtein1966binary}, and suggest words in increasing rates of distance, until a maximum distance of~5. The top 10 words are provided to the user, and if one of these proposed words is selected by the user, the knowledge base and the prediction model are both updated. In our setting, the knowledge base is updated to storage every 2 minutes.

If our system successfully identifies a correctly spelled word, it adds it to the knowledge base directly. Otherwise, the offline model is actively tuned with the newly coming sample. Also, if the user types in a misspelling deliberately and is successfully detected by our system, the system is tuned using that example in the same procedure, and update knowledge base as well if the correct form is provided by the user. 

\section{Methodology}
\label{sec: prediction model}
We test two different types of models to predict if an unseen word is correct or misspelled.  The first uses a LSTM neural model and the second uses a simple character trigram language model.

\subsection{LSTM Model}
We employ a multi-layer Recurrent Neural Network (RNN)~\cite{Rumelhart1986LearningRB} model with Long-Short-Term-Memory cells (LSTM)~\cite{Hochreiter1997LongSM}.
Formally, a sequence $c_1c_2\ldots c_n$ of input characters is first embedded to construct a sequence of vectors $\mathbf{x}_1,\mathbf{x}_2,\ldots,\mathbf{x}_n$.
These are fed trough an LSTM, and the last hidden state $\mathbf{h}_{n}$ is used to make a prediction $\hat{y}$, after passed through a fully connected layer:
\begin{align*}
    \mathbf{x}_i &= \embed(c_i) \\
    \mathbf{h}_t &= \lstm(\mathbf{x_t}; \mathbf{h}_{t-1}) \\
    \hat{y} &= \softmax(\mlp(\mathbf{h}_n)).
\end{align*}

\paragraph{Confusion set}
Neural models require large amounts of training data, hence in our extremely limited resource setting we opt to augment the existing data, building a confusion set for each token in the training set. Each confusion set contains a fixed ratio of the received token (assumed to be correct unless identified by the user) and the rest being misspellings that are fairly similar to the target token in the training set.
The misspellings are generated in following scheme: for each word to be misspelled, we randomly sample an action from the action set of edit distance \{\textit{replace}, \textit{transpose}, \textit{insert}, \textit{delete}\} with equal probability. Then we randomly choose the position and character to conduct this action. Although there is no hard limit for the maximum time of editions to generate another valid word or reasonable misspellings, we find that with more than single time of edition, it is hard even for human users to correct such misspellings. Therefore we only apply a single edit action in misspelling generation.

Moreover, we incorporate linguistic knowledge into misspelling generation. To be specific, we employ several language specific patterns, including vowel switching and doubling/replacing specific consonants in the token. Even though phonetic similar words could contribute in a better training~\cite{Toutanova2002PronunciationMF, li2018spelling}, it is hard to scale to low-resource languages if little phonetic knowledge is given to build a pronunciation model. Thus, we build a character-level confusion set with half of them being generated misspellings.

\paragraph{Data Augmentation}
The augmented training set for the neural approach can be constructed using the confusion sets with three different schemes: 
\begin{enumerate}
    \item \texttt{table}: using the table of most frequent seed tokens only (henceforth \texttt{LSTM-table}),
    \item \texttt{freq}: extending the table such that the vocabulary follows its distribution in the real corpora, so that the least frequent word only appears once in the augmented corpora (henceforth \texttt{LSTM-freq}), where the frequencies are added with $\epsilon = 10^{-5}$ for smoothing purpose.
    \item \texttt{logFreq}: the table extended with number of words follows the log value of its real appearance in the original corpora with least value being 1 (henceforth \texttt{LSTM-logFreq}).
\end{enumerate} 

The two latter approaches incorporate word frequency information in the model by using the training sets produced by the \texttt{logFreq} and \texttt{freq} augmentation approaches. In this way, the training data matches the expected distributional properties of what we expect the realistic use-case scenarios to be. The main difference, between the two, however, is crucial. The proportion of rare words when using the log probabilities (\texttt{logFreq}) is enhanced, which should help with handling more rare words.

\subsection {Character Trigram Model}

We also apply a character-level trigram language model (\texttt{CharTriLM}). The likelihood is calculated in terms of log probabilities, and the likelihood of the word being correct is the sum of log probabilities from all its trigrams. To be specific, for word $w$ of length $n$, $c_{i}$ denotes the character at the $i$-th position of the word, the likelihood $\mathcal{L}(w)$ is calculated in Equation~\ref{eq: loglikelihood}, where $\langle S \rangle$ and $\langle E \rangle$ are the start and ending symbol of the word respectively, $C(s)$ counts the appearances of $s$ in the training data, and add-one smoothing is applied to calculate conditional probabilities in Equation~\ref{eq: smoothing}. Length normalization is employed by averaging likelihoods (Equation~\ref{eq: loglikelihood}) to leverage bias towards short tokens. A threshold bound is then employed to make the decision, where words with log-likelihood lower than the threshold are considered misspelled.

\begin{align}
    P(w) = & P(\langle S \rangle \langle S \rangle c_{0})P(c_{1}|\langle S \rangle c_{0})\notag \\
    & (\prod_{i=2}^{n}P(c_{i}|c_{i-2}c_{i-1}))P(\langle E \rangle |c_{n-2}c_{n-1}) \notag
\end{align}

\begin{align}
    \label{eq: smoothing}
    P(c_{i}|c_{i-2}c_{i-1}) = \frac{C(c_{i-2}c_{i-1}c_{i}) + 1}{\sum_{c}(C(c_{i-2}c_{i-1}c) + 1)}
\end{align}

\begin{align}
    \label{eq: loglikelihood}
    \mathcal{L}(w) = \frac{1}{n}\log P(w)
\end{align}

\section{Datasets}
\label{sec: datasets}
We first benchmark our models on English, using the TOEFL11 dataset~\cite{Blanchard2013TOEFL11AC} which contains 12,100 essays from speakers of 11-non English native languages who took the TOEFL test in 2006-2007. It consists of 1.28m correctly spelled words and 30k misspellings.
Another realistic resource is the Russian dataset for spelling correction~\cite{sorokin2016spellrueval}, which contains 2,000 sentences and the corresponding corrections, containing 23k tokens in total. 

For our synthetic misspellings experiments, we use data from Wikipedia pages of specific topics (native language, famous politicians, football clubs, landmarks) as our resource for Finnish, Italian, Spanish, Turkish and Russian. We select the most frequent words as our training set (seed set).

We split a development set and a test set of 200 words respectively for both the real and synthetic errors datasets. The development set is randomly sampled from the corpus selecting words that are not the~500 most frequent ones, which allows for some vocabulary overlap as more training data are introduced. The test set similarly is sampled from the vocabulary excluding the~1000 most frequent words. The reason we apply this strategy is because we want to mimic the real scenario where incremental information is received from the users, so we need to make sure there is no leakage of test data into the training process -- besides, the most frequent words are also typically the shortest ones and hence the least often misspelled.

\begin{figure*}[t]
    \centering
    \begin{tabular}{cc}
    \includegraphics[width=.5\textwidth]{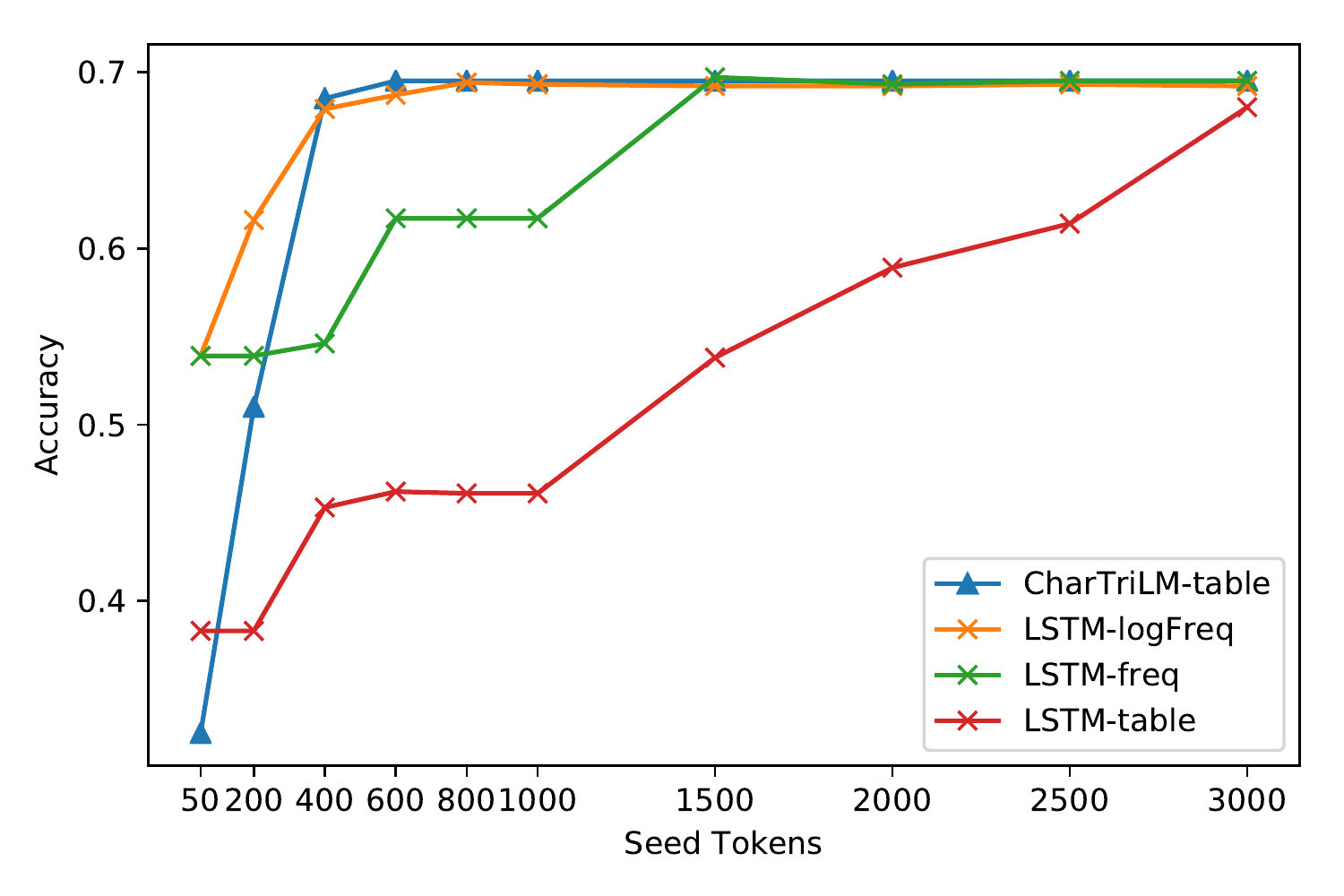} & \includegraphics[width=.5\textwidth]{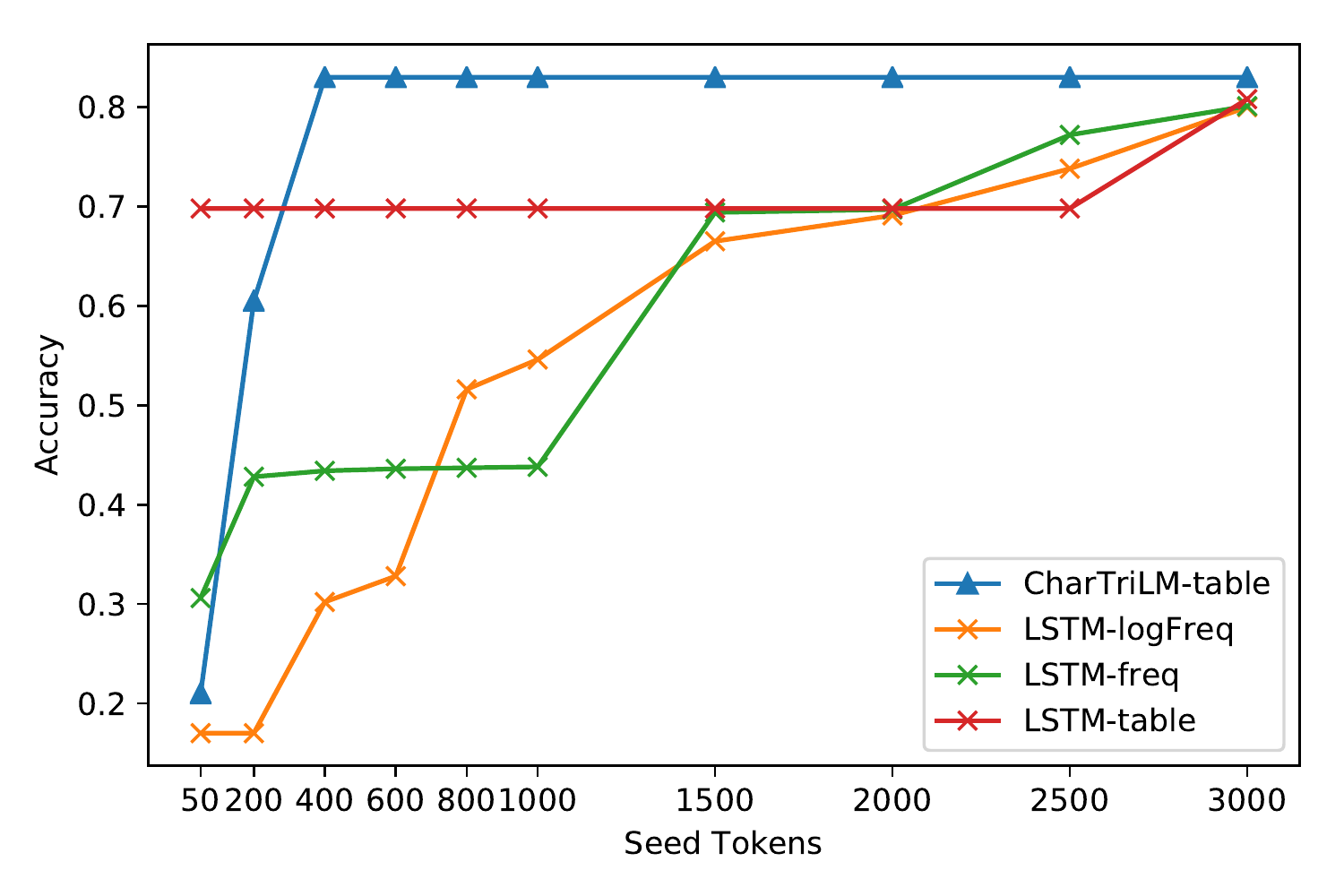}\\
    (a) Accuracy on English & (b) Accuracy on Russian \\
    \end{tabular}
    \caption{Accuracy of different models on datasets with real misspellings. In English (a) \texttt{CharTriLM} and \texttt{LSTM-logFreq} perform comparably. Trigram character language models provide better convergence and performance with small data size for Russian (b), while the neural LSTM models only produce similar accuracy when provided with larger training size (which is proportional to the number of seed tokens).}
    \label{fig: model-selection}
\end{figure*}

We also test our model on OCR outputs on two low-resource languages: Ainu and Griko. Ainu is a critically endangered language spoken in the northern Hokkaido island in Japan, and our Ainu data are the result of a currently-underway digitization project of the book \textit{``Yukara no kenkyu Ainu Jojishi"}, which includes transcriptions of Ainu epic poems (\textit{yukar}) using Latin characters, along with Japanese translations. Griko is an endangered Greek dialect spoken in south Italy, written in the Latin alphabet, and we use data from a digitization project \cite{anastasopoulos+al:coling2018} of the \textit{``Racconti greci inediti di Sternatia,''} a book of traditional Griko stories collected by Prof. Paolo Stomeo in the mid-1900s. We calculate most frequent words based on separate monolingual data (3069 vocabulary from 51k tokens in Griko and 12k vocabulary from 201k tokens in Ainu), train the model on corpora extended from the frequent-word table, and evaluate on the OCR outputs. Note, however, that the monolingual data follows different spelling conventions than the OCR-ed data (the digitized books are from the early 20\textsuperscript{th} century; the monolingual data follows quite different 21\textsuperscript{st} century conventions).

Just like Griko, many low-resource or endangered languages lack a standardized transcription/orthographic standard, and we envision that our system will eventually lead to a better understanding of what the \textit{speakers of the language} use and/or need, as our spelling correction models will be purely based on user-provided data. As a result, they will implicitly lead to organically-produced spelling normalization/standardization.

\section{Experiments and Results}
\subsection{Setup}
\label{sec: setup}

For the neural models, all words are padded with \textit{$\langle PAD \rangle$} to the max token length of 16, and we add \textit{$\langle SOS \rangle$} and \textit{$\langle EOS \rangle$} at the token's beginning and end. We set the number of LSTM layers to 3, hidden LSTM state size in LSTM to 30 and character embedding size to 50. We employ Adam optimizer, the initial learning rate is set as 0.0001 and a batch size of 15. Dropout rate of the fully connected layer is 0.1.\footnote{We performed a limited grid search over parameter settings in our preliminary experiments.} We choose a seed set from the the most frequent words. The size of confusion set is 15 for each token, and misspelling rate is 0.5. We train the neural models for a single epoch, since the lack of data on our envisioned scenarios means that we will not have access to a prior development set.\footnote{We also tested the model by employing an early stopping strategy, but saw no explicit benefits.} 
The model is trained in steps where the most frequent tokens as a seed token set is trained firstly, and then the less frequent ones, the split is made by 50 and 100, and 100 in each the following steps. 

For the trigram character language model, the only parameter to tune is the threshold, which is chosen by finding the threshold that maximizes the F1 score over the training data. We identify the threshold be performing grid search at interval of 0.05 from the minimum to the maximum among training data log-likelihoods. As more seed tokens become available, we simply update the counts and probabilities to incorporate the new data and the pick a new threshold.

\subsection{Model Comparison on Natural Data}
\label{sec: model selection}

We first conduct experiments on real spelling mistakes on the available English and Russian corpora, which allow us to compare the different settings and models.

Results with the trigram character language model, as well as with the LSTM model using all data augmentation schemes are shown in Figure~\ref{fig: model-selection}. 
Interestingly, results seem to vary depending on the language. On English (Figure~\ref{fig: model-selection}(a)) the \texttt{LSTM-logFreq} and the \texttt{CharTriLM} perform comparably as more data are added to the model. On the other hand, in the Russian dataset (Figure~\ref{fig: model-selection}(b)) the character trigram language model almost consistently outperforms the neural approach, which only achieves similar performance when more training data are available.

We believe that this difference is mainly due to the inconsistencies of English spelling, which lead the ``simpler" character trigram model astray. It is worth noting that the \texttt{logFreq} approach works much better for English. We suspect that frequency-based data augmentation information can indeed be more useful, but that a development set is necessary in order for it to be effective. In our realistic uses-case scenarios, a gold-annotated development set will not be available from the beginning, but will become available shortly after the system receives consistent feedback from multiple community members/users, which will also allow the system to make better frequency estimates.

For the rest of our experiments, therefore, we choose our two best models, \texttt{CharTriLM} trained on the vocabulary table and LSTM trained on set extended by log word frequency (\texttt{LSTM-logFreq}), for all our comparisons.

\begin{figure}[t]
    \centering
  \includegraphics[width=0.48\textwidth]{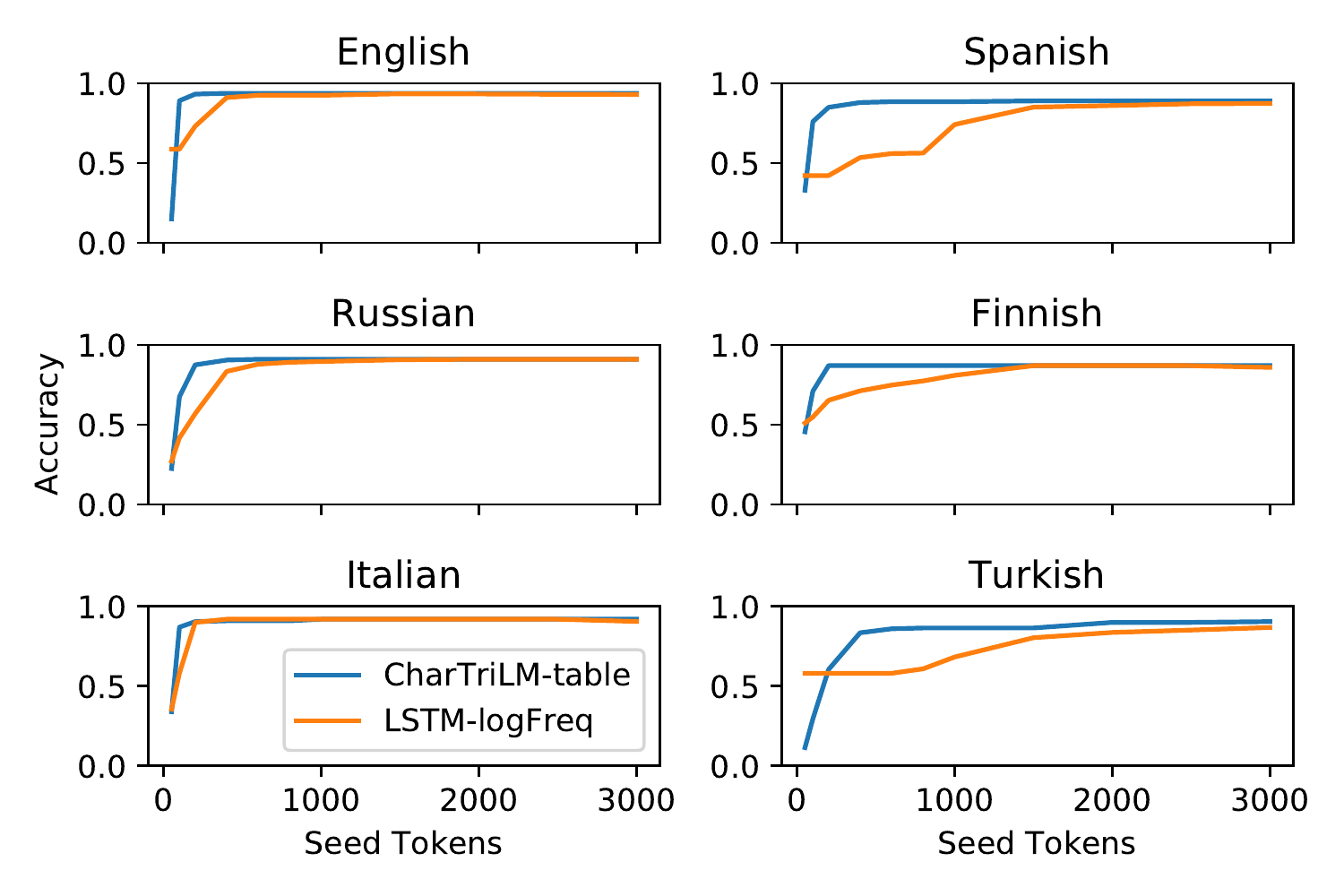} 
  \caption{Accuracy on synthetic data. \texttt{CharTriLM} consistently outperforms \texttt{LSTM} in the extremely low-resource cases.}
  \label{fig: multilingual performance} 
\end{figure}

\subsection{Evaluation on Synthetic Data}
\label{sec: data synthesis}

In addition to the natural spelling mistakes from the previous section, we are interested in studying the performance of our model on more languages with varying degrees of morphological complexity.
We synthesize errors on Wikipedia articles on English, Russian, Italian, Spanish, Finnish, and Turkish, as described in Section~\ref{sec: datasets}, and train and test our two best models (\texttt{CharTriLM} and \texttt{LSTM-logFreq}) on these synthetic data.

The results are shown in Figure~\ref{fig: multilingual performance}. The trends on all languages are very similar: the character trigram LM is better in the beginning when fewer seed tokens are available for training, but the neural approach is able to quickly match that performance. The exact point differs per language e.g. for Italian is as low as~400 tokens, while for Spanish and Turkish it is around~1500 to~2000 tokens.

An overview comparison of different languages with training data restricted to 500 most frequent words is in Table~\ref{tab:500-training-result}. The trigram character language model can achieve an accuracy of more than 60\% from this small set of tokens, without exact frequency information included. \texttt{LSTM-logFreq} achieves comparable performance in some cases (e.g. English, Finnish) or even slightly higher (Italian) but generally lags behind.

\begin{table}[ht]
    \small
    \centering
    \begin{tabular}{c|cc|cc}
     \toprule
        \multirow{2}{*}{Corpora} &
        \multicolumn{2}{c|}{\texttt{LSTM-logFreq}} &
        \multicolumn{2}{c}{\texttt{CharTriLM}}\\
        \cmidrule{2-5}
        & Accuracy & F1 & Accuracy & F1 \\
    \midrule
        \multicolumn{1}{l}{\textit{Real data}} \\
         English & \textbf{0.696} & 0.452 & 0.635 & \textbf{0.455} \\
        Russian & 0.301 & 0.217 & \textbf{0.830} & \textbf{0.454} \\
    \midrule
        \multicolumn{1}{l}{\textit{Synthetic data}} \\
        English & 0.925 & \textbf{0.488} & \textbf{0.938} & 0.484\\
        Russian & 0.254 & 0.161 & \textbf{0.910} & \textbf{0.476} \\
        Italian & \textbf{0.920} & \textbf{0.479} & 0.910 & 0.476 \\
        Spanish & 0.561 & 0.330 & \textbf{0.880} & \textbf{0.468} \\
        Finnish & 0.729 & 0.435 & \textbf{0.870} & \textbf{0.465}\\
        Turkish & 0.581 & 0.320  & \textbf{0.845} & \textbf{0.458} \\
    \midrule
    \multicolumn{1}{l}{\textit{OCR outputs}}\\
    Griko & 0.562 & 0.369 & \textbf{0.590} & \textbf{0.467} \\
        Ainu & 0.746 & 0.420 & \textbf{0.950} & \textbf{0.487} \\
    \bottomrule
    \end{tabular}
    \caption{Performance of neural and character-trigram language models trained on 500 most frequent words. \texttt{CharTriLM} shows better performance in most corpora, although in afew cases the \texttt{LSTM-logFreq} model achieves comparable performance.}
\label{tab:500-training-result}
\end{table}

\subsection{Results on Endangered Language OCR Outputs}

Last, we investigate the performance of our spelling correction models on Ainu and Griko. It is worth pointing out that the spelling correction model now has to deal with a different type of errors: optical character recognition produces different misspelling patterns, based on the characters' shape, rather than  their perceived pronunciations or typing device. For example, $\chi$ in the Griko documents is often mis-recognised as \textit{x} and $\grave{c}$ is often wrongly recognised as \textit{d}. Thus the confusion set could only been constructed through random selection of actions without morphological knowledge. 

\begin{figure}[t]
  \centering
  \includegraphics[width=0.48\textwidth]{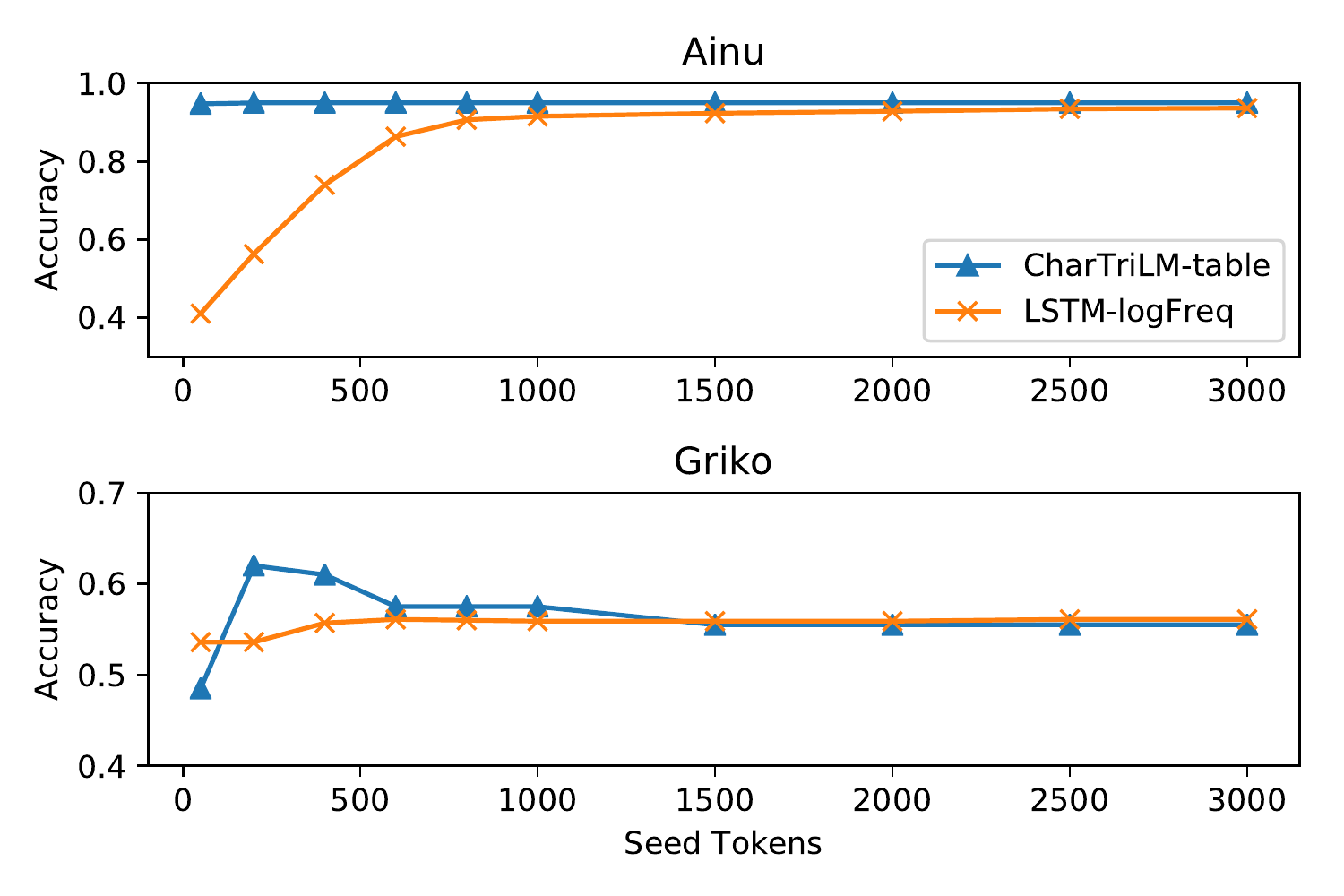} 
  \caption{Performance on OCR output data for Ainu and Griko. The difference in performance is due to different Griko spelling conventions between training and testing data.}
  \label{fig: ocr performance} 
\end{figure}

Nevertheless, our models are still able to acquire information gain from the seed tokens. The accuracy of the models' is shown in Figure~\ref{fig: ocr performance}. Again, the patterns are quite different depending on the language. 

On one hand, accuracy on the Ainu dataset is quickly over~90\% with the \texttt{CharTriLM} model and the LSTM model is able to achieve similarly high accuracy with less than~1000 seed tokens. This is because the quality of the OCR outputs is in fact very high for the Ainu dataset, leading to highly unbalanced real datasets -- in fact, more than~90\% of the tokens are indeed correct.

On the other hand, the performance of both models on the Griko data hovers around~50\% regardless of how much additional seed tokens we use for augmentation and training. This is partly due to the OCR output being of much lower quality than the Ainu one, with a Character Error Rate of more than 40\% (which leads to even higher Word Error Rate, meaning that most words are actually incorrect).
In addition, though, this behaviour is further exacerbated because of different spelling conventions between the monolingual data used to provide the seed tokens, and the spelling conventions of the digitized Griko text. As a result, using additional data from the actual data as they get corrected/annotated would be a lot more beneficial than using more monolingual data.
Although in our envisioned real use-case scenario our system will also function as a spelling standardization/normalization system on top of a spelling correction one, as more users add more data to be used for training, the goal of OCR post-correction is \textit{not} to normalize the spelling, but rather accurately convey the original transcribed text.

\section{Online Case Study}

We have not yet performed large-scale user experiments on multiple languages, but we have tested the system with one targeted user: the Hindi/English speaker was asked to write a 500 word paragraph about his favorite movie, using romanized Hindi plus English (Hinglish). 

We setup online testing with neural model and character trigram language model as back-end models. The manual correction is conducted at the end of each sentence. The experiment is conducted both in the scenario with pre-trained model and without any pre-trained models loaded (i.e. without any knowledge of the target language), and no knowledge base is stored for all the settings. Word frequencies were calculated from articles collected from Hinglishpedia.\footnote{https://www.hinglishpedia.com/}
The corpus contains 89k tokens with a vocabulary of 5,856 types, out of which we select the top 300 most frequent words as the training set for pre-training. Both systems performed reasonably for the user, who was not informed of the difference between settings. Surprisingly, the user commented that the setting using the LSTM system seemed to be better at generalization, mentioning specifically that it got (Hindi) plurals correct more often than the other system. 

We plan to conduct further case studies on multiple languages in order to further solidify our findings regarding the suitability of our approach for low-resource languages. 

We will also freely release our code and our prototype system, which should allow for any interested community members to use it. We also plan to convert our prototype into a plugin for popular messaging/texting applications that allow such modifications, in the hope of encouraging more users to text and communicate in their native languages.

\section{Conclusion}
Spelling normalization is hard to acquire for low-resource languages when nearly no labelled misspelling data is available. In this work, we propose a system with a simple prediction model to aid spelling normalization. We designed an user interface to employ interactive strategies and gain correct/misspelling labels from the user. Experimental results shows that the abnormal spelling patterns can successfully be identified even with only a small set of most frequent correct words by a character-level language model.  However the difference between the character level language model and the LSTM model becomes minimal when a reasonable amount of data is available.  We are aware that neural language models make better use of large amounts of data, though are often brittle when there is only fewer training examples, thus a hybrid system with character languages models for the very low resource used initially, followed by the LSTM system when more data becomes available seems the best compromise.

\bibliography{emnlp2019}
\bibliographystyle{acl_natbib}

\end{document}